\documentclass[conference]{IEEEtran}
\usepackage[latin9]{inputenc}
\usepackage{array}
\usepackage{float}
\usepackage{url}
\usepackage{multirow}
\usepackage{amsmath}
\usepackage{amssymb}
\usepackage{graphicx}
\usepackage{xcolor}
\usepackage{soul}
\usepackage{flushend}
\usepackage[unicode=true, bookmarks=false, breaklinks=false,pdfborder={0 0 1},backref=section,pdftex,pdfborder={0 0 0},colorlinks=true,allcolors=darkblue]{hyperref}
\definecolor{darkblue}{rgb}{0.0, 0.0, 0.55}

\makeatletter
\DeclareMathOperator*{\argmin}{arg\,min}
\providecommand{\tabularnewline}{\\}

\definecolor{Mem}{rgb}{.75,1,0}
\sethlcolor{Mem}

\makeatother

\begin{document}
% paper title

\title{Deep Knowledge Tracing and Dynamic Student Classification for Knowledge Tracing}

\author{\IEEEauthorblockN{Sein Minn$^{1}$, Yi Yu$^{2}$, Michel C. Desmarais$^{1}$, Feida Zhu $^{3}$, Jill-Jênn Vie$^{4}$}
\\
\IEEEauthorblockA{
$^{1}$Department of Computer Engineering, Polytechnique Montreal, Canada \\
$^{2}$Digital Content and Media Sciences Research Division, National Institute of Informatics, Japan \\
$^{3}$School of Information Systems, Singapore Management University, Singapore \\
$^{4}$RIKEN Center of Advanced Intelligence Project, Japan \\
}}
\maketitle
\begin{abstract}
In Intelligent Tutoring System (ITS), tracing the student's knowledge state during learning has been studied for several decades in order to provide more supportive learning instructions. In this paper, we propose a novel model for knowledge tracing that i) captures students' learning ability and dynamically assigns students into distinct groups with similar ability at regular time intervals, and ii) combines this information with a Recurrent Neural Network architecture known as Deep Knowledge Tracing. Experimental results confirm that the proposed model is significantly better at predicting student performance than well known state-of-the-art techniques for student modelling.
\end{abstract}


\footnote{This work is available at \url{https://github.com/simon-tan/DKT-DSC.git}}
\section{Introduction}

\label{sec:intro}

%%%%%%%%%%%%%%%%%%%%%%%%%%%%%%%%%%%%%%%%%%%%%%%%%%%%%%%%%%%%%%%%%%%%%%%%%%%%%%%

ITS is an active field of research that aims to provide
personalized instructions to students. Early work dates back to the late 1970s. A wide array of Artificial Intelligence and Knowledge Representation techniques have
been explored, of which we can mention rule-based and Bayesian representation of student
knowledge and misconceptions, skills modeling with logistic regression in Item Response
Theory, case-based reasoning, and, more recently reinforcement learning
and deep learning \cite{brown1978diagnostic,polson2013foundations}.  One can even argue that
most of the main techniques found in Artificial Intelligence and Data Mining have found
their way into the field of ITS, and in particular for the problem of knowledge tracing,
which aims to model the student's state of mastery of conceptual or procedural knowledge
from observed performance on tasks \cite{corbett1994knowledge}.

% The task of learning a student model \Mem{learning student models or modeling student's skills acquisition?} and predicting how students learn \Mem{we predict how student perform, not how students learn} is known as an interdisciplinary research topic across education, psychology, neuroscience and cognitive science. One of the main challenges for ITS designers is tracing student's knowledge and learning behaviour for providing more supportive pedagogical instruction adaptively. As human learning is grounded with complexity in various dimensions such as human brain, knowledge, experiences and practices, it is inherently difficult to trace student's knowledge.

% Knowledge tracing is a key component of ITS, the task of learning a student model is to predict whether students will get correct answers for next problems during their interactions with ITS. According to student's performance, different teaching instructions and suggestions can be recommended for their next step in the system, individually and dynamically. Specifically, the model will decide what kind of problems with associated skills have been mastered by students and choose the problems for practice in their next step based on their performance. Decisions about skipping or delaying suggestions should also be automatically updated for each individual student.

In this paper we propose a novel model for knowledge tracing, Deep Knowledge Tracing with
Dynamic Student Classification (DKT-DSC).  At each time interval, the model first assigns a student into a distinct group
of students that share similar learning ability. This information is then fed to a
Recurrent Neural Network (RNN), known as the DKT architecture \cite{piech2015deep} for predicting
student's performance from data.  We can consider the student classification as a long-term
memory of the student's ability as input to the RNN improves
knowledge tracing with DKT, which is among the state-of-the-art approach to knowledge tracing.

The rest of this paper is organized as follows. Section~\ref{sec:relatdwork}
reviews related work on student modelling techniques. Section~\ref{sec:IDKT} presents
the proposed DKT-DSC model.
Section~\ref{sec:dataset} describes the datasets used in our experiments.
Experimental results are shown in Section~\ref{sec:experimental-study}
and finally Section~\ref{sec:conclusion--future} concludes this
paper and discusses future avenues of research.

\section{Related Work}\label{sec:relatdwork}

%% The goal of student modelling is to measure students' proficiency based on their previous
%% interactions with the system and trace their knowledge state and learning behaviour. Most of
%% the current Intelligent tutoring systems need human instructions to find appropriate content
%% that fit for student's current knowledge state.  Popular educational platforms such as
%% Coursera, EdX require student models for better understanding student learning styles by
%% using large scale student interaction data to model students and improve their online
%% educational experiences.

We review here four of the best known state-of-the-art student modelling methods for
estimating student's performance, either for their predominance in psychometrics (IRT) or Educational Data
Mining (BKT), or because they are best performers (PFA, DKT). See \cite{desmarais2012review} for a general review.

\subsection{Item Response Theory (IRT) }
\label{sec:IRT}

IRT assumes the student knowledge state is static and represented by her proficiency when completing an assessment during an exam~\cite{wilson2016back,van2013handbook,gonzalez2014general,ekanadham2017t}. IRT models a single skill and assumes
the test items are \textit{unidimensional}.  It assigns student $i$ with a static proficiency $\theta_{i}$. Each item $j$ has its own difficulty $\beta_{j}$.
The main idea of IRT is estimating a probability that student $i$
answers item $j$ correctly by using student's ability and item's
difficulty. The widely used one-parameter version of IRT, known as the Rasch model, is
\begingroup\makeatletter\def\f@size{9}\check@mathfonts 

\begin{equation}
p_j({\theta_i})=\frac{1}{1+e^{-({\theta_i}-\beta_j)}}.
\end{equation}
\endgroup
Recently, Wilson \cite{wilson2016back} proposed an IRT model
that outperforms state-of-the-art knowledge tracing models. In which, maximum a posteriori (MAP) estimates of $\theta_{i}$ and $\beta_{j}$ are computed using the Newton-Raphson method.

\subsection{Bayesian Knowledge Tracing (BKT)}

\label{sec:BKT}

BKT  was introduced for knowledge tracing within a learning environment for which the assumption on static knowledge states is dropped~\cite{corbett1994knowledge,d2008more}.  It also assumes a single skill is tested per item, but this assumption is relaxed in later work on BKT.
Standard BKT estimate of student's knowledge about
a skill is continually updated with four probabilities:
{[}$P(L_{0})$ initial probability of mastery, $P(T)$ transitioning from non-mastery to mastery,
$P(G)$ guessing and $P(S)$ slipping{]}, once the student gives her response
at each time:

\begingroup\makeatletter\def\f@size{8}\check@mathfonts
\begin{equation} \label{equ:corr}
P(L_{n}|Correct) = \frac{P(L_{n-1})(1-P(S))}{P(L_{n-1})(1-P(S))+(1-P(L_{n-1}))P(G)}
\end{equation}

\begin{equation} \label{equ:incorr}
P(L_{n}|Incorrect) = \frac{P(L_{n-1}) P(S)}{P(L_{n-1})P(S)+(1-P(L_{n-1}))(1-P(G))}
\end{equation}

\begin{equation} \label{equ:action}
P(L_{n}) = P(L_{n-1}|Action)+ (1-P(L_{n-1}|Action))P(T)
\end{equation}
\endgroup
There have been various extensions of BKT in the last decades \cite{Baker2008More,pardos2011kt}.
% FIXME: add reference

\subsection{Performance Factor Analysis (PFA)}
\label{sec:PFA}

PFA, which was proposed as an alternative to BKT, also relaxes the static knowledge assumption and models multiple skills simultaneously~\cite{pavlik2009performance} with its basic structure. It defines the probability of success to an item~$j$ by student~$i$ as:
\begingroup\makeatletter\def\f@size{9}\check@mathfonts 
\begin{equation}
P(m_{i,j}) = 1 / (1 + e^{-\ell_{i,j}})
\end{equation}
\begin{equation}
\ell_{i,j} = \beta_{j}+\sum_{k\in KCs}(\gamma_{k}s_{ik}+\rho_{k}f_{ik}),\label{equ:pfm}
\end{equation}
\endgroup
% FIXME: there was a collision here, so I replaced m with the logit \ell
where $\beta_{k}$ is the bias for the skill $k$, and $\gamma_{k}$
and $\rho_{k}$ represent the learning gain per success and failure attempt
to skill $k$, respectively. $S_{ik}$ is the number of successful attempts and
$f_{ik}$ is the number of failure attempts made by student $i$ on skill $k$~\cite{pavlik2009performance}.

\subsection{Deep Knowledge Tracing (DKT)}
\label{sec:DKT} 

DKT was introduced in \cite{piech2015deep}.  It uses a Long
Short-Term Memory (LSTM)\cite{hochreiter1997long} to represent the latent
knowledge space of students dynamically. The increase in student's knowledge through an assignment can be inferred by utilizing the history of student's previous performance. DKT uses large numbers of artificial
neurons for representing latent knowledge state along with a temporal
dynamic structure and allows a model to learn the latent knowledge
state from data. It is defined by the following equations:
\begingroup\makeatletter\def\f@size{10}\check@mathfonts 
\begin{equation}
h_{t}=\tanh(W_{hx}x_{t}+W_{hh}h_{t-1}+b_{h}),\label{equ:hidden}
\end{equation}
\begin{equation}
y_{t}=\sigma(W_{yh}h_{t}+b_{y}).\label{equ:output}
\end{equation}
\endgroup
In DKT, both tanh and the sigmoid function are applied element wise
and parameterized by an input weight matrix $W_{hx}$, recurrent weight
matrix $W_{hh}$, initial state $h_{0}$, and readout weight matrix
$W_{yh}$. Biases for latent and readout units are represented by
$b_{h}$ and $b_{y}$.

\section{Deep Knowledge Tracing with Dynamic Student Classification}
\label{sec:IDKT}

Human learning is a process that involves practice: we become proficient through practice.  However, learning is also affected by the individual's ability to learn, or to become proficient with more or less practice.  We refer to the ability to become proficient with little practice as the learning ability. Based on that notion, we proposed a model Deep Knowledge Tracing with Dynamic Student Classification (DKT-DSC), that assesses a student's
learning ability and assign her into a distinct group of students with similar
ability, and then the model invokes an RNN to trace her knowledge in each distinct group at different
time intervals. It can
trace the performance of students based on their learning ability, reassessed regularly over time.

\subsection{Dynamic assessment of student's learning ability and grouping}\label{sec:DC}

Dividing students into distinct groups with similar learning ability, according
to their previous performance on various contents in a learning system,
has been explored in several research works in the field of education
\cite{merceron2005clustering,trivedi2011clustering} for providing more adaptive instructions to
each group of students with similar ability.  Dynamic assessment of student learning ability at each time interval is performed
by clustering based on the assessment of their previous performance
history before the start of next time interval. 

\subsubsection{Time interval}  Time interval is a segment containing a number of student's attempts to answer questions in the system. In this perspective, a tick of time is a single first attempt to a question or exercise. 

\subsubsection{Segmenting students' attempt sequence}

\label{sec:seg}

segmentation of each student response sequence into multiple time intervals serves two purposes:
1) To reduce computational burden and memory space allocation for learning
throughout a long sequence. 
2) To re-assess a student' learning ability after each time interval and assign
her into a group which she belongs to for the next time interval dynamically. 

\begin{figure}[H]
\centering \includegraphics[width=8.5cm]{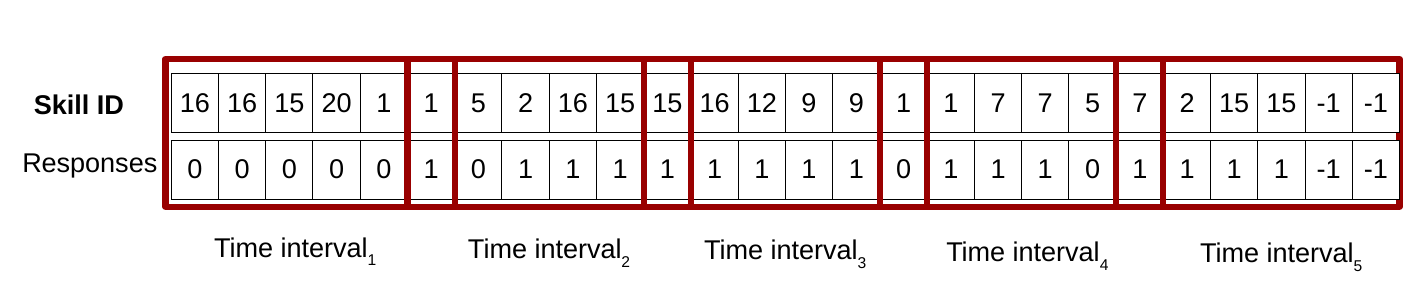} \caption{Segmentation of a student's attempt sequence.}
\label{fig:seg} 
\end{figure}

Fig. \ref{fig:seg} illustrates an example of dividing a 24-attempt response
sequence of a student into 5 segments (time intervals)
where a segment represents a time interval in which that student answered
6 problems in the system. When the student stopped interacting with
system, it is represented with -1 in the last time interval. The number
of attempts made by each student varies based on the number of questions
they answered during the interaction with system.  

\subsubsection{Long-term skills encoding for clustering}

\label{sec:dataextract} Student are grouped according to their learning ability profile: the skills or knowledge they acquired. Data for assessing student's learning ability is available from previous attempts on test items or exercises corresponding to a specific 
skill. 

The learning ability profile is encoded as a vector of length the number of skills, and updated after each time interval
by using all previous attempts on each skill. The differences between success and failure ratios on each skill of
student's previous attempts are transformed into a data vector for
clustering student $i$ at time interval $z$ as follows: 
\begingroup\makeatletter\def\f@size{9}\check@mathfonts
\begin{equation}
Correct(x_{j})_{1:z}=\sum_{t=1}^{z}\frac{(x_{jt}==1)}{|N_{jt}|},\label{equ:segcorr}
\end{equation}
% FIXME: suggestion: what about replacing $x_{jt} == 1$ with $S_{jt}$? It would make the connection with the number of successes in PFA.

\begin{equation}
Incorrect(x_{j})_{1:z}=\sum_{t=1}^{z}\frac{(x_{jt}==0)}{|N_{jt}|},\label{equ:segincorr}
\end{equation}

\begin{equation}
R(x_{j})_{1:z}=Correct(x_{j})_{1:z}-Incorrect(x_{j})_{1:z},\label{equ:ratio}
\end{equation}

\begin{equation}
d_{1:z}^{i}=(R(x_{1})_{1:z},R(x_{2})_{1:z},...,R(x_{n})_{1:z}),\label{equ:seg}
\end{equation}
\endgroup
in which $Correct(x_{j})_{1:z}$ and~$Incorrect(x_{j})_{1:z}$ represent the ratios of skill~$x_{j}$
being correctly answered or incorrectly, by student~$i$ on~$n$ number of skills
$(x_{1},x_{2},..,x_{n})$ from time interval 1 to current time interval~$z$. $|N_{jt}|$is the total
number of practices of skill~$x_{j}$ up to time interval $t$. $R(x_{j})_{1:z}$ represents the difference
between how much student~$i$ performs on skill~$j$, correctly or
incorrectly, for time interval~$1$ to $z$ and~$d_{1:z}^{i}\in D$ represents
a vector containing the learning ability profile of student~$i$ on
each skill from time interval 1 until~$z$. Each student may have a different number of total time intervals in the lifetime of their interactions with the system (see Fig.~\ref{fig:cluster}).

\subsubsection{K-means Clustering}
\label{sec:clustering}

\begin{figure}[H]
	\centering \includegraphics[width=8.5cm]{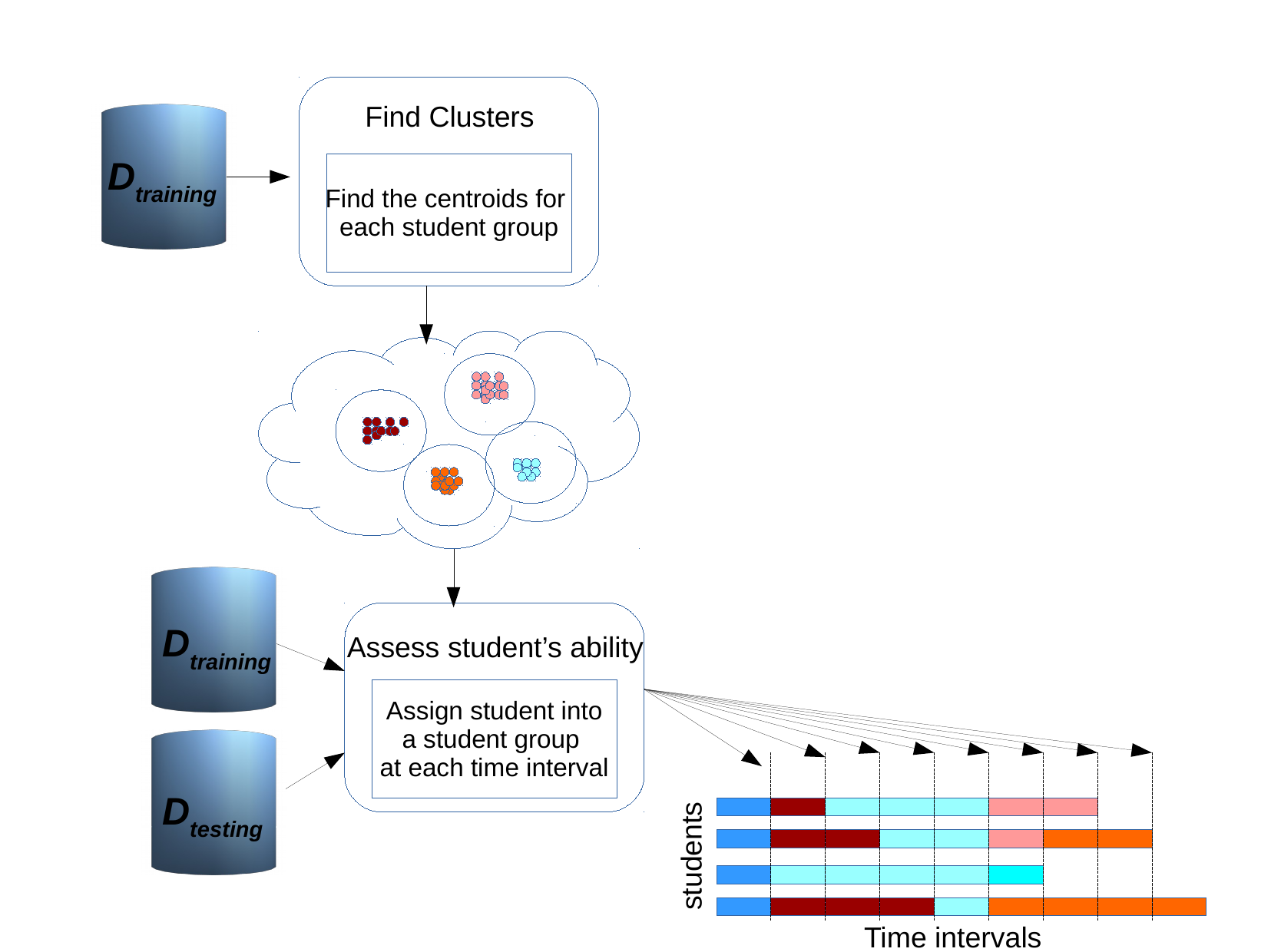} \caption{Clustering students at each time interval.}
	\label{fig:clustering} 
\end{figure}

Assigning students into a group with similar ability
at each time interval is performed by k-means clustering on data~$D$ \cite{macqueen1967some,ball1965novel}. At the time of the clustering training phase, we find the centroids for each student group without considering the time interval index. 
Once it has been computed, the centroid of each group will not change
any more during the whole clustering process. After that, we assign students (in both training and testing data)
into distinct groups at each time interval (see Fig \ref{fig:clustering}).
When we find the group which student~$i$ belongs to at time interval
$z$, we use the learning ability profile data points~$d_{1:z-1}^{i}$
because we are not supposed to know the current attempts of student~$i$ at time interval~$z$. After learning the centroids   of all $K$ clusters, each student at each time interval
$Seg_{z}$ is assigned into the nearest cluster~$C_{c}$ by the following equation:
\begingroup\makeatletter\def\f@size{9}\check@mathfonts
\begin{equation}\label{equ:cluster}
Cluster(Stu_{i},Seg_{z}) = \argmin_{C} \sum^K_{c=1} \sum_{d^i_{1:z-1}\in C_c} ||d^i_{1:z-1}- \mu_c||^2
\end{equation}
\endgroup
where~$\mu_{c}$ is the mean of points in a cluster set~$C_{c}$ (a
group of students), and ability profile data~$d_{1:z-1}^{i}$ represents the previous performance data of student~$i$ from time interval 1 to~$z-1$.

\begin{figure}[H]
\centering \includegraphics[width=8.5cm]{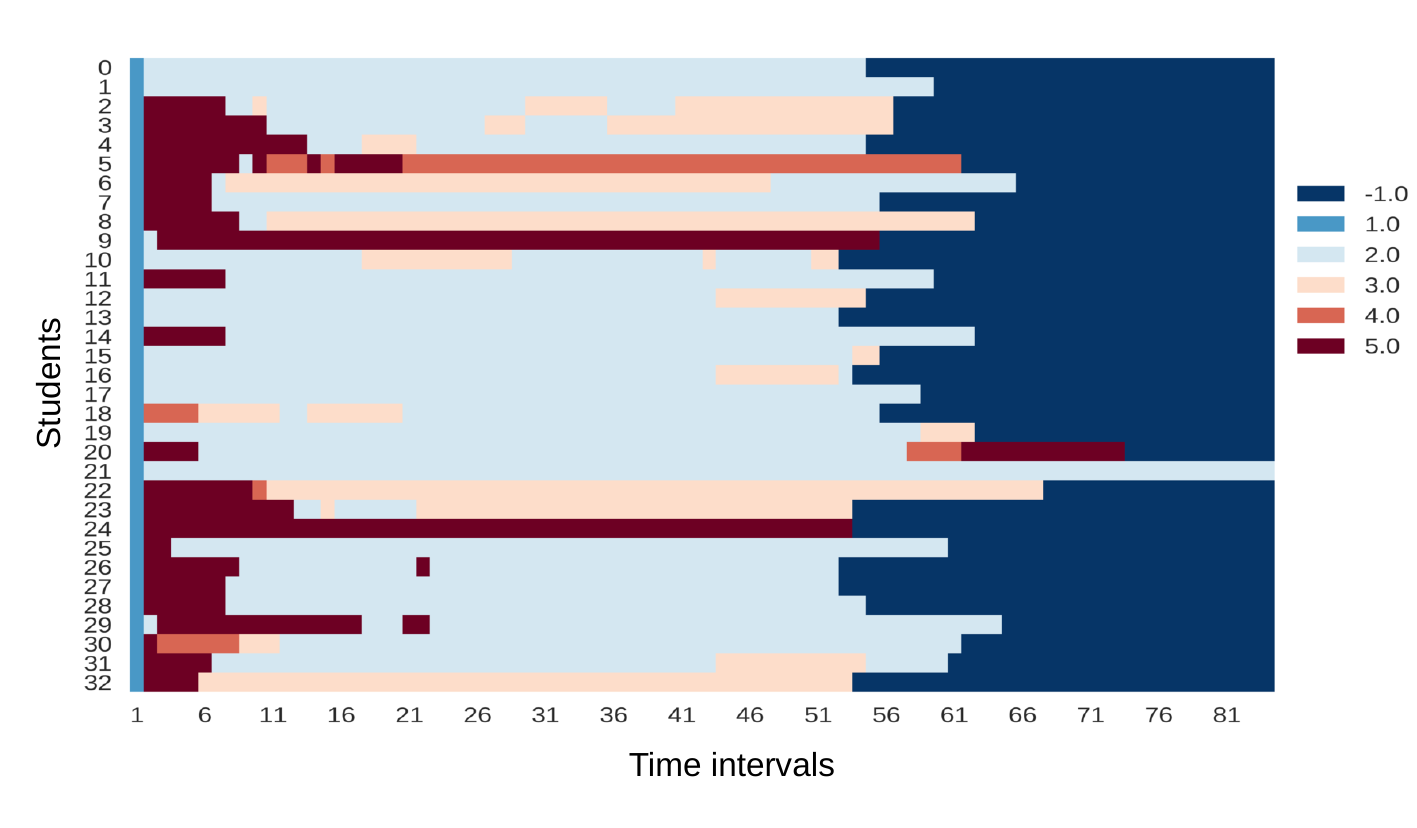} \caption{Evolution of students' learning ability over each time interval (each time interval contains 20 attempts) throughout
their interactions.}
\label{fig:cluster} 
\end{figure}

Figure \ref{fig:cluster} illustrates the data of 33 students' learning abilities
based on their previous performance and the evolution over time intervals. Dark blue (-1) means students do not have any attempt
by the time when they quit the system. Group 1 is for
the first time interval of every student and the rest of the groups~$(2,..,5)$
are assigned by the k-means clustering method at each time interval
$z$ by using previous performance data~$d_{1:z-1}^{i}$.

\subsection{Deep knowledge tracing}
\label{sec:DKT-DSC}

DKT-DSC incorporates student's learning ability
to the DKT for better individualization of the system, by assigning
a student into a group of students with similar ability dynamically.
It relaxes the assumption that all students have the same ability
and that students' ability is consistent over time. In fact, student's
ability is evolving continuously and some students may learn faster
than others. 

In the standard DKT,~$x_{t}$ is a one-hot encoding vector of the
student interaction tuple $x_{t}=\{s_{t},a_{t}\}$ that represents
the combination of the skills~$s_{t}$ practiced, and of~$a_{t}$ which indicates
if the answer is correct. But DKT-DSC requires~$x_{t}=\{s_{t},a_{t}\}$ additionally
with~$c_{t}$ which is a group or cluster~$Cluster(Stu_{i},Seg_{z})$
indicating~$Stu_{i}$'s ability at current time interval~$Seg_{z}$ . In the hidden
layers, the last node of each time interval is served as first node~$h_{0}$
for next time interval when we segment the response sequence into
multiple time intervals. The output~$y_{t}$ is a vector of same length
as the number of problems. Thus, the probability of the next problem
answered correctly at $c_{t}$ of $Seg_{z}$  can be obtained from~$y_{t}$. In that respect,
Eq.~\ref{equ:hidden} and~\ref{equ:output} are still valid for DKT-DSC. The output~$y_{t}$
of both DKT and DKT-DSC is the same which provides the predicted probability
for a particular problem. 
\begin{figure}[H]
\centering \includegraphics[width=9cm]{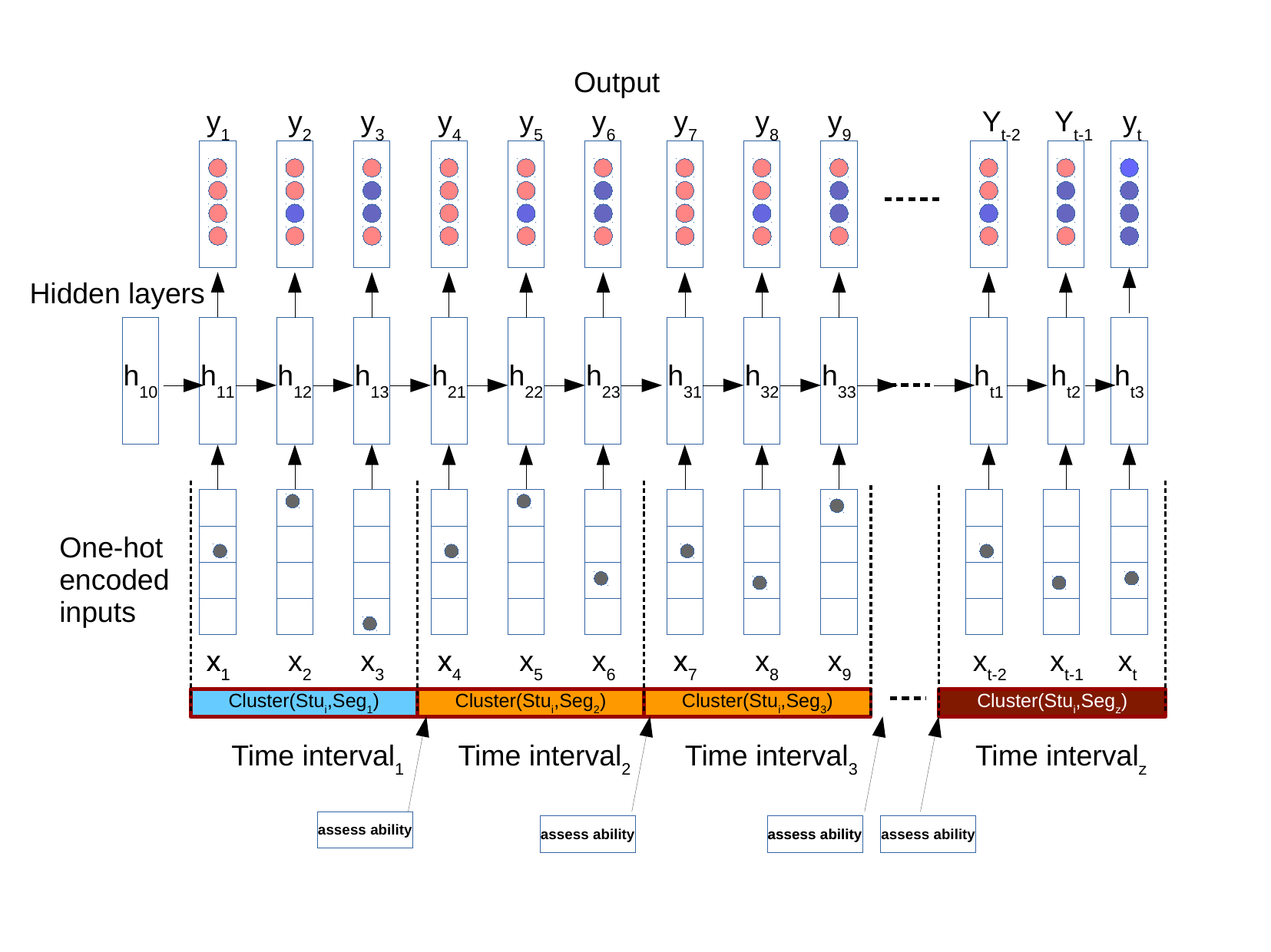} \caption{DKT-DSC prediction in each time interval (each segment) is associated
with a distinct group (cluster) throughout interactions of a student
with the system.}
\label{fig:IDKT} 
\end{figure}

Figure \ref{fig:IDKT} illustrates how DKT-DSC model has been adapted
by incorporating student's learning ability as distinct group information at
each time interval (each segment) to improve individualization
in knowledge tracing. The colour at each time interval at the input layer
% FIXME: "at the input layer" -> Should it be "in the input layer"?
represents which group a student belongs to at that time interval according to her learning ability.  Note that without incorporating student's ability,
DKT-DSC model is the same as the standard DKT model.

By adding this cluster information $Cluster(Stu_{i},Seg_{z})$ of what group the student belongs to, we ensure that these high-level statistics are still available to the model for making its predictions throughout the whole academic year. This is what the DKT model does, treating all students in same way without considering their learning abilities.  On the contrary, DKT-DSC uses clustering to find a group of students with similar ability by using their ability profile data at different
time intervals. Tracing student's knowledge in each different group can provide more
effectiveness in student's performance prediction.

Finally, we summarize the characteristics of each model in this paper
in Table \ref{tab:comparison}.

\begin{table}[H]
	\caption{Comparison of different models}
	
	\label{tab:comparison} %  \vskip -0.3cm
	\begin{centering}
		\begin{tabular}{|c|c|c|c|c|c|}
			\hline 
			& IRT  & PFA  & BKT  & DKT  & DKT-DSC\tabularnewline
			\hline 
			Use of student's ability & Yes  & No  & No  & No  & Yes\tabularnewline
			Use of item difficulty & Yes  & No  & No  & No  & No \tabularnewline
			Use of single skill & Yes  & No  & Yes  & Yes  & Yes \tabularnewline
			Use of multiple skill & No  & Yes  & No  & No  & No \tabularnewline
			Learn on ordered sequence & No  & No  & Yes  & Yes  & Yes \tabularnewline
			\hline 
		\end{tabular}
		\par\end{centering}
	%\vskip -0.4cm
	
	\vspace{-2mm}
\end{table}

\section{Datasets}

\label{sec:dataset}

In order to validate the proposed model, we tested
it on four public datasets from two distinct tutoring scenarios in
which students interact with a computer-based learning system in
educational settings.

\begin{itemize}
	\item The ASSISTment system\footnote{\url{https://sites.google.com/site/assistmentsdata}}
is an online tutoring system that was first created in 2004 which
engages middle and high-school students with scaffolded hints in their
math problems. If students working on ASSISTments answer a problem
correctly, they are given a new problem. If they answer it incorrectly,
they are provided with a small tutoring session where they must answer
a few questions that break the problem down into steps. Datasets are: ASSISTments 2009-2010 skill builder data set, ASSISTments 2012-2013,	ASSISTments 2014-2015. 

In all datasets, problems are usually tagged with just one skill,
but a rare few may be associated with two or three skills. It typically
depends on the structure given by the content creator. Some researchers separate a record with multiple skills into multiple single skill records by duplicating. Wilson
\cite{wilson2016back} claimed that this type of
data processing can artificially boost prediction results significantly,
because these duplicate rows can be accounted for approximately 25\%
of the records in the Assistment09 dataset for DKT models. So we
removed duplicate and multiple-skill repeated records in all datasets for the fairness of comparison.
% FIXME: I feel this part is unclear. You don't actually describe what kind of data processing, and the fact the duplicate rows were existing is a different thing.
\item KDD Cup: The PSLC DataShop released several data sets derived
from Carnegie Learning's Cognitive Tutor. Algebra 2005-2006 \cite{corbett2001cognitive}
is a development dataset released during the KDD Cup 2010 competition\footnote{\url{https://pslcdatashop.web.cmu.edu/KDDCup/downloads.jsp}}.
In this dataset, the problems are associated with multiple skills. So we regard a subset
of multiple skills as a new skill \cite{xiong2016going}.
\end{itemize}

We evaluate models described above with four datasets from two separate
real world tutors. The experimental results show how the models perform
across different datasets. Only the first correct attempts
to original problems are considered in our experiment.

To the best of our knowledge, these are the largest publicly available
knowledge tracing datasets.

\begin{table}[H]
\caption{Overview of datasets}

\label{tab:data} %  \vskip -0.3cm
\begin{centering}
\begin{tabular}{|c|r|r|r|c|}
\hline 
\multirow{2}{*}{Dataset} & \multicolumn{3}{c|}{Number of} & \multirow{2}{*}{Description}\tabularnewline
\cline{2-4} 
 & \multicolumn{1}{c|}{Skills}  & \multicolumn{1}{c|}{Students} & \multicolumn{1}{c|}{Records} & \tabularnewline
\hline 
\cline{1-1} 
\multirow{3}{*}{ASSISTments} & 123  & 4,163  & 278,607  & 2009-2010  \cite{razzaq2005assistment} \tabularnewline
 & 198  & 28,834  & 2,506,769  & 2012-2013  \cite{feng2009addressing} \tabularnewline
 & 100  & 19,840  & 683,801  & 2014-2015 \cite{xiong2016going} \tabularnewline
\hline 

Cognitive Tutor  & 437  & 574  & 808,775  & KDD Cup 2010 \cite{corbett2001cognitive}
\tabularnewline
\hline
\end{tabular}
\par\end{centering}
%\vskip -0.4cm

\vspace{-2mm}
\end{table}

\section{Experimental Study}

\label{sec:experimental-study}

DKT-DSC is extended from the original DKT algorithm, and is
combined with the k-means clustering method with a Euclidean distance. Ten iterations were made in the training stage. DKT-DSC and DKT share the same
loss function, with 200 fully-connected
hidden nodes for each hidden layer. For speeding up the training process,
mini-batch stochastic gradient descent is used to minimize the loss
function. The batch size for our implementation is 32, corresponding to 32 split
sequences from each student. We train the model with a learning
rate of 0.01 and dropout is also applied for avoiding overfitting \cite{srivastava2014dropout}.

In our experiment, 5 fold cross-validations are used to make predictions
on both datasets. Each fold involves randomly splitting each dataset
into 80\% training data and 20\% test data at the student level. So
both training and test datasets contain response records from different
students. Training for clustering is performed only using data from
students in the training dataset. We use EM to train BKT and the limit
of iterations is set to 200. We learn models for each skill and make
predictions separately, then the results for each skill are averaged.
For DKT and DKT-DSC, we set the number of epochs to 100. All these models
are trained and tested on the same sets of data. Next response of
a student is predicted by using current and previous response sequence in chronological order.

We compare our model with state-of-the-art models: IRT \cite{wilson2016back}, BKT \cite{Baker2008More}, PFA \cite{pavlik2009performance},
DKT \cite{piech2015deep}. But we do not compare with other
variant models, because those are more or less similar and do not
show significant difference in performance. For IRT, we
apply the code from Knewton \cite{wilson2016back} and the code for
DKT is from WPI \cite{xiong2016going}. For DKT, we use the same setting
of parameters as DKT-DSC and also apply segmentation for a fair comparison.
Predicted sequences of student performance by each model are tabulated
and evaluated in terms of Area Under the Curve (AUC) and Root mean
squared error (RMSE). AUC provides a robust metric
where the value to predict is binary, as it is the case of our datasets. An AUC of 0.50 represents
the score achieved by random guess. We set AUC 0.61 of BKT as
a baseline in our experiment. 

\begin{table}[h]
	\begin{center}
		\caption{AUC result for all datasets}\label{tab:exp1}
		\begin{tabular}{|cccccc|}
			\hline
			
			\multirow{2}{*}{Datasets} & \multicolumn{5}{c|}{Model} \\
			\cline{2-6}
			& BKT & IRT& PFA &  DKT & DKT-DSC      \\			
			\hline
			ASSISTments09  & 0.651 & \textbf{0.751}&0.703 & 0.721& \textbf{0.735}  \\  
			\hline
			ASSISTments12  & 0.623 & \textbf{0.743}& 0.670& 0.713& \textbf{0.721}   \\  
			\hline
			ASSISTments14  & 0.611 & 0.672& 0.689& 0.707 & \textbf{0.716}   \\  
			\hline
			Cognitive Tutor & 0.642 & \textbf{0.806}& 0.760& 0.784 & \textbf{0.792}    \\
			\hline

		\end{tabular}  
	\end{center}	
\end{table}

In Table \ref{tab:exp1}, DKT-DSC performs better than
state-of-the-art models in all datasets. On the ASSISTments09 dataset,
compared with the standard DKT which has an AUC of 0.721, our DKT-DSC
model achieves an AUC of 0.735, which represents a significant gain of 2\%. On the ASSISTments12 dataset with 2.5
million records, the result shows an increase, AUC 0.721 in DKT-DSC
compared with AUC 0.13 in the original DKT. In the latest ASSISTments14
dataset, DKT-DSC achieves an improvement of 1\% over the original DKT.
In the Cognitive Tutor dataset, DKT-DSC also achieves about 1\% gain
with AUC=0.792 while the original DKT has AUC=0.784. As for other algorithms,
IRT also provides a slight improvement over the original DKT in all
datasets but DKT-DSC performs significantly better than both DKT and
IRT. Note that Problem ID is not provided in the original ASSISTments14
dataset. So we use Skill ID as Problem ID for the IRT model, and that
is why IRT only gets a AUC of 0.67. In all models described above, only
the IRT model learns the problem difficulty while all other models
only rely on skills.

\begin{table}[h]
	\begin{center}
		\caption{RMSE result for all datasets }\label{tab:exp2}
		\begin{tabular}{|cccccc|}
			\hline
			
			\multirow{2}{*}{Datasets} &  \multicolumn{5}{c|}{Model} \\
			\cline{2-6}
			& BKT &IRT& PFA & DKT &DKT-DSC   \\
			
			ASSISTments09  & 0.471 & 0.440 & 0.454 & 0.45 & \textbf{0.430}  \\  
			\hline
			ASSISTments12 & 0.510  & 0.441 & 0.44& 0.45 & \textbf{0.431}  \\  
			\hline
			ASSISTments14  & 0.510 &  0.441 & 0.42 & 0.429 & \textbf{0.430}  \\  
			\hline
			Cognitive Tutor & 0.44 & 0.374 & 0.391&  \textbf{0.38} & \textbf{0.373}   \\
			\hline

		\end{tabular}  
	\end{center}	
\end{table}

In Table \ref{tab:exp2}, when we compare the models in terms of RMSE,
BKT is 0.46 in ASSISTments09, 0.51 in ASSISTments12 and 0.47 in Cognitive
Tutor. RMSE results of DKT-DSC in all dataset are under 0.43 while that
of all other models are no less than 0.44 (except IRT, PFA and DKT
in the Cognitive Tutor dataset). According to these results, DKT-DSC
outperforms in all ASSISTments datasets and shows a slightly better performance in
the Cognitive Tutor dataset.
% FIXME: "makes a vast majority of statistical differences" does not mean anything
All of the above experiments are conducted on the time interval containing 20 attempts and 8 clusters (groups of students).

\section{Conclusion and Future Work}

\label{sec:conclusion--future} In this paper, we proposed a new model,
DKT-DSC that assesses student's learning ability at each time interval
and  dynamically assigns a student into a distinct group of students with the same
ability. A student's knowledge is traced based on the group
which she belongs to at each time interval. Experiments with four
datasets show that the proposed model performs statistically
and significantly better than state-of-the-art models. DKT assumes
all students have the same learning ability and only tracks the improvement
of knowledge in a skill sequence without considering difference between
abilities of each student and learning rate. In comparison, our model
improves over DKT by capturing the student's ability over time. Assessing student's ability in this
way gives the model critical information in the prediction of student
performance in their next time interval and tracing their knowledge
where abilities of the students evolve dynamically. We individualize the
input vector by taking both student's ability and practicing skill
into account. Instead of using the skill level alone, incorporating student's
ability in terms of group information in DKT-DSC yields an improvement
in the prediction of performance. Dynamically assessing student's ability at each
time interval plays the critical role and helps the DKT-DSC model capture
more variance in the data, leading to more accurate predictions.

In our future work, we will adapt this model to problems with
multiple associated subskills in the system and apply it in the recommendation
of problems with multiple associated skills. Problems
for practice should be recommended according to the knowledge level
and the ability a student possesses. The significant gain obtained
by DKT-DSC can make a difference in current knowledge tracing respect.
Further investigation on the potential application of DKT-DSC to other
content recommendations (movies and other commercial products) will
also be considered.

%%%%%%%%%%%%%%%%%%%%%%%%%%%%%%%%%%%%%%%%%%%%%%%%%%%%%%%%%%%%%%%%%%%%%%%%%%%%%
\section{NOTE:}
the results are different than the published ICDM version. A bug was found and corrected.  While the scale of the differences is lower, the general trends are identical.  The major difference is that DKT-DSC performed better than IRT* in the published results, whereas it does not in the corrected results.
\section{Acknowledgements}

This work was partially done while the first author was interning at
National Institute of Informatics, Japan and also funded by the NSERC
Discovery funding awarded to the third author.

\bibliographystyle{IEEEtran}
\bibliography{IEEEabrv,biblio}

\end{document}